\documentclass{article}

 \usepackage[preprint]{neurips_2025}

\usepackage[utf8]{inputenc} 
\usepackage[T1]{fontenc}    
\usepackage{hyperref}       
\usepackage{url}            
\usepackage{booktabs}       
\usepackage{amsfonts}       
\usepackage{nicefrac}       
\usepackage{microtype}      

\usepackage{amsmath}
\usepackage{graphicx}
\usepackage{subcaption}
\usepackage{listings}
\usepackage[dvipsnames]{xcolor}

\usepackage{wrapfig}

\newcommand{\ours}{$\text{SaPE}^2$}
\newcommand{\oursfull}{2-Dimensional Semantic-Aware Position Encoding}
\title{A 2D Semantic-Aware Position Encoding for Vision Transformers}

\author{
\begin{minipage}[t]{0.95\textwidth}
\centering
Xi Chen\textsuperscript{1}\thanks{Equal Contribution.}, 
Shiyang Zhou\textsuperscript{1}\footnotemark[1], 
Muqi Huang\textsuperscript{2}, 
Jiaxu Feng\textsuperscript{3}, 
Yun Xiong\textsuperscript{1}\thanks{Corresponding Author.},
Kun Zhou\textsuperscript{2}, \\
Biao Yang\textsuperscript{1}, 
Yuhui Zhang\textsuperscript{1}, 
Huishuai Bao\textsuperscript{1}, 
Sijia Peng\textsuperscript{1},
Chuan Li\textsuperscript{2}, 
Feng Shi\textsuperscript{2} \\
\vspace{1mm}
\textsuperscript{1}Shanghai Key Laboratory of Data Science, School of Computer Science, \\ Fudan University, Shanghai, China \\
\textsuperscript{2}Alibaba Group, Shanghai, China \\
\textsuperscript{3}Department of Physics, Fudan University, Shanghai, China \\
\vspace{1mm}
\texttt{\scriptsize \{x\_chen21, 22210240079\}@m.fudan.edu.cn, huangmugi.hmg@alibaba-inc.com, feng.jiaxu@outlook.com,} \\
\texttt{\scriptsize yunx@fudan.edu.cn, kun.zhouk@alibaba-inc.com, \{22210240343, 23212010049, 22212010001\}@m.fudan.edu.cn,} \\
\texttt{\scriptsize pengsijia1216@gmail.com, \{lc357677, sam.sf\}@alibaba-inc.com}
\end{minipage}
}

\begin{document}

\maketitle

\begin{abstract}

Vision transformers have demonstrated significant advantages in computer vision tasks due to their ability to capture long-range dependencies and contextual relationships through self-attention. However, existing position encoding techniques, which are largely borrowed from natural language processing, fail to effectively capture semantic-aware positional relationships between image patches. Traditional approaches like absolute position encoding and relative position encoding primarily focus on 1D linear position relationship, often neglecting the semantic similarity between distant yet contextually related patches. These limitations hinder model generalization, translation equivariance, and the ability to effectively handle repetitive or structured patterns in images.
In this paper, we propose \oursfull~(\ours), a novel position encoding method with semantic awareness that dynamically adapts position representations by leveraging local content instead of fixed linear position relationship or spatial coordinates. Our method enhances the model’s ability to generalize across varying image resolutions and scales, improves translation equivariance, and better aggregates features for visually similar but spatially distant patches. By integrating \ours~into vision transformers, we bridge the gap between position encoding and perceptual similarity, thereby improving performance on computer vision tasks.
\end{abstract}

\section{Introduction}
\label{sec:intro}

Transformer has been widely adopted in both natural language processing (NLP) \cite{transformer} and computer vision (CV) \cite{vit, swin} due to its superior ability to capture long-range dependencies, efficient parallel computation, and flexible contextual modeling. Compared to traditional convolutional neural networks (CNNs) \cite{o2015introduction, li2021survey}, transformer-based architectures exhibit greater flexibility, enabling seamless adaptation to complex tasks such as cross-modal learning (e.g., joint image-text processing), image generation, and high-level semantic reasoning. Their self-attention mechanism provides strong expressive power, making them highly competitive in visual understanding.

Unlike CNNs \cite{o2015introduction}, which operate on spatially structured grid-based feature maps, vision transformers process images in a sequence-based manner \cite{vit}. Typically, an input image is partitioned into a set of non-overlapping patches, which are then linearly embedded into tokens before being fed into a transformer model \cite{vit, swin, detr}. These tokens are treated equally and interact through self-attention, where pairwise attention scores determine their relationships. However, since the self-attention mechanism itself is position-agnostic (permutation-invariance of self-attention operation), explicit position encoding (PE) is required to introduce location information into the model. To this end, position encodings are integrated into the model’s attention mechanism, ensuring that the computed attention scores incorporate spatial relationships between tokens.

\begin{figure*}[t]
    \centering
    \includegraphics[width=1\linewidth]{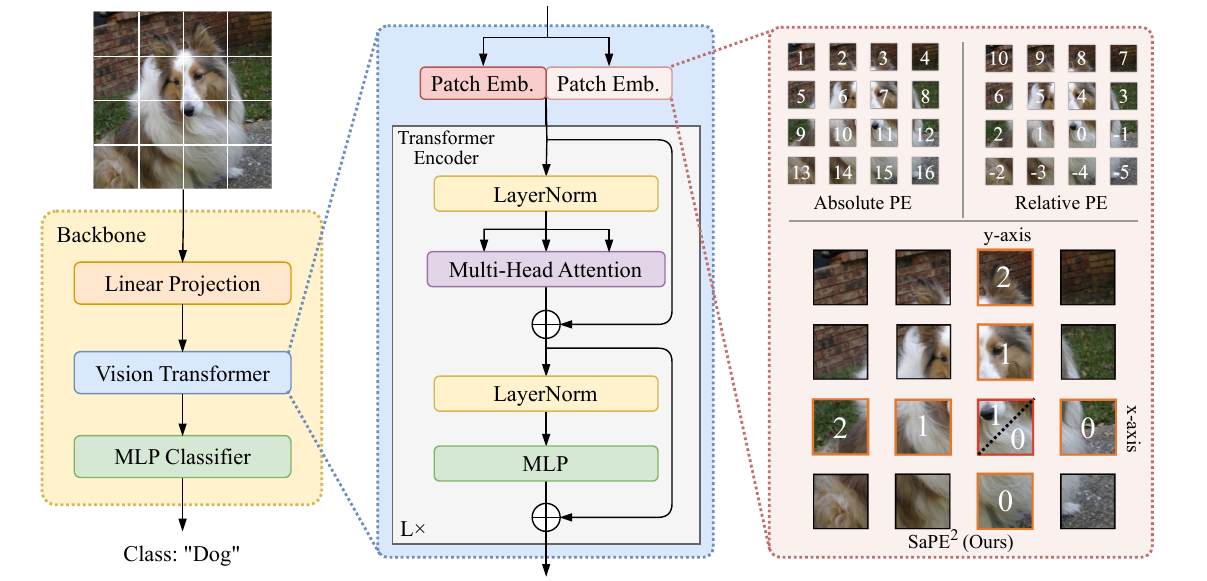}
    \caption{The overall pipeline of the vision model and a comparison between the absolute PE, relative PE, and \ours. The patch with the red border is the target patch for relative PE and our proposed method (\ours). The number in each patch indicates the position information of the patch. This is just a schematic diagram; please refer to Figure \ref{fig:pipeline} for details on \ours.}
    \label{fig:compare}
    \vspace{-15pt}
\end{figure*}

While existing position encoding techniques in computer vision effectively incorporate spatial awareness into vision transformers, they rely exclusively on explicit spatial coordinates while overlooking semantic-aware positional relationships between patches. Specifically, absolute position encoding (APE) \cite{transformer, gehring2017convolutional} is coordinate-dependent, making it sensitive to image resolution and scale variations, thereby limiting its adaptability to different input sizes. Relative position encoding (RPE) \cite{shaw2018self, raffel2020exploring} (including Rotary Position Embedding \cite{rope}) focuses only on spatial displacement, without incorporating the semantic similarity between different regions of the image. None of these methods consider the semantic relationships among patches, meaning they fail to capture perceptual similarity beyond local adjacency.

These limitations are particularly problematic in structured and repetitive patterns, where visually similar regions are treated as completely independent simply due to their spatial separation. 
For instance, two patches representing the same object category (e.g., two instances of a car in an image) might receive distinct encodings based on their absolute coordinates, even though they share strong semantic similarity. This discrepancy prevents the model from effectively recognizing and leveraging their relatedness. Similarly, in images with repetitive textures or grid-like structures, conventional position encodings may fail to exploit the redundancy in attention computation, as semantically correlated tokens are treated as independent, leading to suboptimal information aggregation.

Moreover, position encoding methods that rely solely on coordinate-based distances hinder translation equivariance. When an object shifts within an image, traditional position encoding schemes alter its spatial representation, leading to inconsistencies in feature extraction. This can be particularly detrimental to model robustness in image classification, where maintaining spatially consistent representations is crucial for recognizing objects regardless of their position in the image.

These challenges highlight the need for a more flexible, semantic-aware position encoding approach—one that adapts dynamically to local content rather than relying solely on fixed spatial coordinates.

To address the aforementioned limitations, we propose \oursfull~(\ours), which encodes positional relationships based on semantic similarity rather than predefined spatial distances. By dynamically adjusting position representations according to local context, our method enables vision transformers to (1) generalize more effectively across different resolutions and scales, (2) improve translation equivariance, and (3) enhance feature aggregation for visually similar yet spatially distant regions.

Our proposed semantic-aware position encoding allows the model to assign similar positional representations to semantically related regions, enhancing its understanding of global context. For instance, in the given image in Figure \ref{fig:compare} (a dog resting near a corner wall), traditional position encoding methods would treat each patch independently based solely on spatial coordinates. In contrast, our method recognizes that patches corresponding to the dog belong to a coherent foreground entity, while patches representing the brick wall and grass form the background. By assigning similar position encodings to semantically related patches, our approach helps the model better differentiate between the foreground and background, leading to more consistent feature representations.
By integrating \ours~into vision transformers, our method bridges the gap between previous position encoding and perceptual similarity, leading to improved generalization in downstream tasks.

We summarize our contributions as follows:
\begin{itemize}
    \item We propose a semantic-aware approach to position encoding, integrating it with the inherent 2D spatial structure of image inputs. This enables vision transformers to capture semantic relationships between patches rather than relying solely on fixed spatial coordinates. Moreover, it ensures a cohesive representation that enhances both spatial and semantic understanding in transformer-based vision models.
    \item We conduct extensive experiments, benchmarking our method against traditional position encoding techniques and thoroughly analyzing its effectiveness and impact on model performance.
    \item We identify the limitations of commonly used position encoding methods in transformer-based vision models and empirically demonstrate that our proposed method offers a promising solution.
\end{itemize}

\section{Preliminaries}

\paragraph{Absolute Position Encoding (APE)} To incorporate the sequential order information into ViTs, APE assigns a unique positional representation to each token. Given the token embeddings $\mathbf{x}_i \in \mathbb{R}^{d_x}$, the absolute position encoding $\mathbf{p}_i \in \mathbb{R}^{d_x}$ is added directly to the embeddings, yielding the final input representation:
\begin{equation}
\mathbf{x}_i' = \mathbf{x}_i + \mathbf{p}_i.
\end{equation}
The position encoding $\mathbf{p}_i$ can be either fixed or learnable. Learnable encodings associate each position $i$ with a unique trainable parameter vector $\mathbf{e}[i]$ ~\cite{vit}, while fixed encodings often leverage sinusoidal functions with varying frequencies ~\cite{transformer}. The sinusoidal position encoding defines each dimension of the position vector as:
\begin{equation} 
\begin{split}
    \mathbf{p}_i^{(2j)} = \sin\left(\frac{i}{10000^{\frac{2j}{d_x}}}\right), \mathbf{p}_i^{(2j+1)} = \cos\left(\frac{i}{10000^{\frac{2j}{d_x}}}\right),
\end{split}
\end{equation}
where $i$ denotes the position index, $j$ is the dimension index, and $d_x$ represents the dimension of the token embeddings. The frequency of the sinusoidal functions is determined by the denominator, which scales the position index $i$ by an exponential function of the dimension index. This design enables the model to generalize to longer sequences.

\paragraph{Relative Position Encoding (RPE)}
Unlike APE, which assigns absolute positional information to each token, RPE models the pairwise distance between tokens, making the encoding invariant to translation. For a pair of tokens at positions $i$ and $j$, the relative position encoding is represented as $r_{i,j}$, and the final attention score is modified as:
\begin{equation}
a_{ij} = \frac{\mathbf{q}_i^\top \mathbf{k}_j + r_{i,j}}{\sqrt{d_k}},
\end{equation}
where $\mathbf{q}_i, \mathbf{k}_j \in \mathbb{R}^{d_k}$ are the query and key representations, and $d_k$ is the dimensionality of the key vectors. The relative position encoding $r_{i,j}$ can be implemented as a learnable parameter ~\cite{shaw2018self} or a function of the relative distance $i - j$.

\paragraph{Rotary Position Encoding (RoPE)}
RoPE is a special form of RPE. It encodes positional information by applying a rotation matrix to the query and key embeddings. This method preserves the relative positional relationship between tokens in the inner product space:
\begin{equation}
    \mathbf{q}_n^{\prime}=\mathbf{q}_n e^{i n \theta}, \mathbf{k}_m^{\prime}=\mathbf{k}_m e^{i m \theta},
\end{equation}
\begin{equation}
    \mathbf{A}_{(n, m)}^{\prime}=\operatorname{Re}\left[\mathbf{q}_n^{\prime} \mathbf{k}_m^{\prime *}\right]=\operatorname{Re}\left[\mathbf{q}_n \mathbf{k}_m^* e^{i(n-m) \theta}\right],
\end{equation}
where $n$ and $m$ denote the positions of the corresponding tokens, and $\theta$ represents the angular frequency that varies across different feature dimensions. The varying frequencies $\theta_t$ are defined as:
\begin{equation}
    \theta_t=10000^{-t /\left(d_{\text {head }} / 2\right)}, \text { where } t \in\left\{0,1, \ldots, d_{\text {head }} / 2\right\},
\end{equation}
where $d_{\text{head}}$ is the dimensionality of each attention head. This design assigns higher frequencies to lower-dimensional features and lower frequencies to higher-dimensional features. The exponential scaling ensures that different dimensions encode positional information at different granularities, with lower dimensions capturing finer positional distinctions and higher dimensions capturing coarser positional patterns. 

\paragraph{Contextual Position Encoding (CoPE)}
CoPE ~\cite{cope} is initially proposed in NLP tasks and it measures positions in a context-dependent way, rather than simple token counts. First, a gate value is computed to determine whether a key token should be included in the position measurement:
\begin{equation}
g_{i j}=\sigma\left(\mathbf{q}_i^{\top} \mathbf{k}_j\right),
\end{equation}
where $j<i$ and $\sigma$ denotes the sigmoid function. The gate value ranges from 0 to 1, indicating whether the token is counted or ignored. The position value is then obtained by summing the gate values between the current and target tokens:
\begin{equation}
p_{i j}=\sum_{k=j}^i g_{i k}.
\end{equation}
If all gates are 1, $p_{i j}$ corresponds to the token-based relative position, making CoPE a generalization of relative position encoding. Generally, $p_{ij}$ can be the count of specific words or word types like nouns or numbers, the number of sentences, or other useful concepts. The final PE can be calculated by interpolating trainable position embeddings $\mathbf{e}[p]$.

\begin{figure*}
    \centering
    \includegraphics[width=1\linewidth]{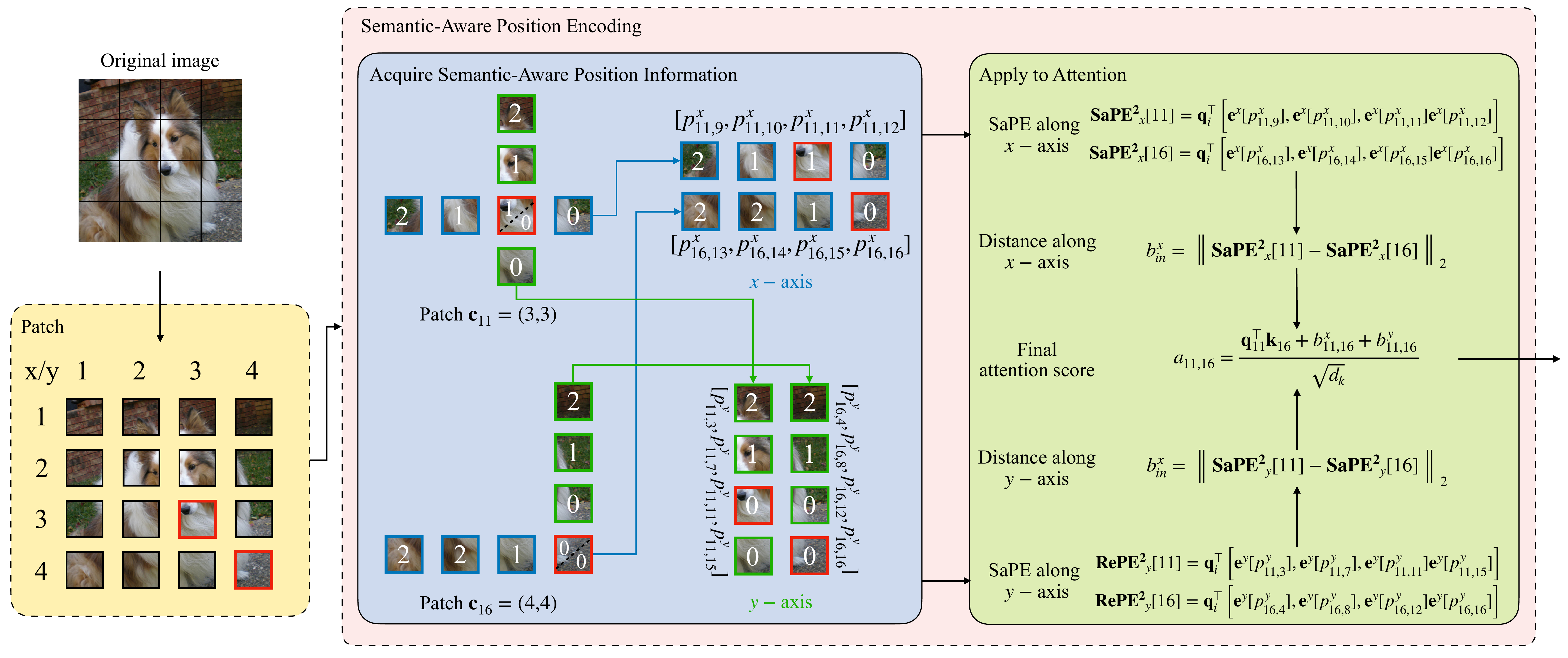}
    \caption{The pipeline of calculating the relative positional relationship between two patches which in the different x-axes and y-axes.}
    \label{fig:pipeline}
\end{figure*}

\section{Method}

In this section, we present the detailed implementation of our proposed \ours~method. The overall architecture is illustrated in Figure~\ref{fig:compare} and \ref{fig:pipeline}.

Given that images are naturally represented as 2D grids, the core idea is to decompose the 2D position encoding into two independent 1D position encodings along the horizontal and vertical axes. This decomposition enables the model to capture spatial information in a separable manner, enhancing both local and global spatial dependencies.

Let $\mathbf{I} \in \mathbb{R}^{H \times W \times C}$ denote an input image, where $H$ and $W$ represent the number of patches along the vertical ($y$) and horizontal ($x$) axes, respectively, and $C$ is the number of channels. The coordinate of the $i$-th patch is denoted by $\mathbf{c}_i = (x_i, y_i)$, where $x_i \in\{1, 2, \ldots, W\}$ and $y_i \in\{1, 2, \ldots, H\}$.

\ours~encodes spatial information by applying independent 1D position encodings along each axis. For the horizontal axis, let $\mathbf{q}_i$, $\mathbf{k}_i$, and $\mathbf{v}_i$ denote the query, key, and value representations of the $i$-th patch. The gate value for each patch pair sharing the same $y$ coordinate is computed as:
\begin{equation}
    g_{ij}^x = \sigma\left(\mathbf{q}_i^{\top} \mathbf{k}_j\right), \quad \text{where} \ y_i = y_j.
\end{equation}
The gate value $g_{ij}^x$ encodes the relative position between the $i$-th and $j$-th patches, enabling the model to perceive spatial relationships such as foreground and background information. The position value along the $x$-axis is obtained by summing the gate values:
\begin{equation}
    p_{im}^x = \sum_{j \in \mathcal{I}_{im}} g_{ij}, \quad \mathcal{I}_{im} = \{ j \mid y_i = y_j = y_m, \ x_m \leq x_j \}.
\end{equation}
Here, $p_{im}^x$ answers the question: what is the relative position of the $m$-th patch from the $i$-th patch's perspective along the $x$-axis? In general, it captures transitions in color, object boundaries, or other latent patterns that the transformer deems useful for understanding the image content.

Since $p_{im}^x$ is not restricted to integer values, we cannot directly convert a position value to a position vector in the same way as the relative PE. Inspired by CoPE \cite{cope}, we use interpolation between integer embeddings to obtain continuous positional representations. Let $\mathbf{e}[p]$ denote the learnable embedding for integer position $p \in \left[0, M\right]$. The interpolated embedding is computed as:
\begin{equation}
\begin{split}
    \mathbf{e}^x\left[p_{im}^x\right]=\left(p_{im}^x-\left\lfloor p_{im}^x\right\rfloor\right) \mathbf{e}^x\left[\left\lceil p_{im}^x\right\rceil\right]+
    \left(1-p_{im}^x+\left\lfloor p_{im}^x \right\rfloor\right) \mathbf{e}^x\left[\left\lfloor p_{im}^x\right\rfloor\right],
\end{split}
\end{equation}
where $\mathbf{e}^x\left[p_{im}^x\right]$ represents the position embedding of the $m$-th patch relative to the $i$-th patch along the $x$-axis. This embedding encodes the spatial relationship between the two patches, answering the question: what does it mean for the $m$-th patch to be at this particular position from the $i$-th patch's perspective?

\begin{figure}
    \centering
    \includegraphics[width=0.5\linewidth]{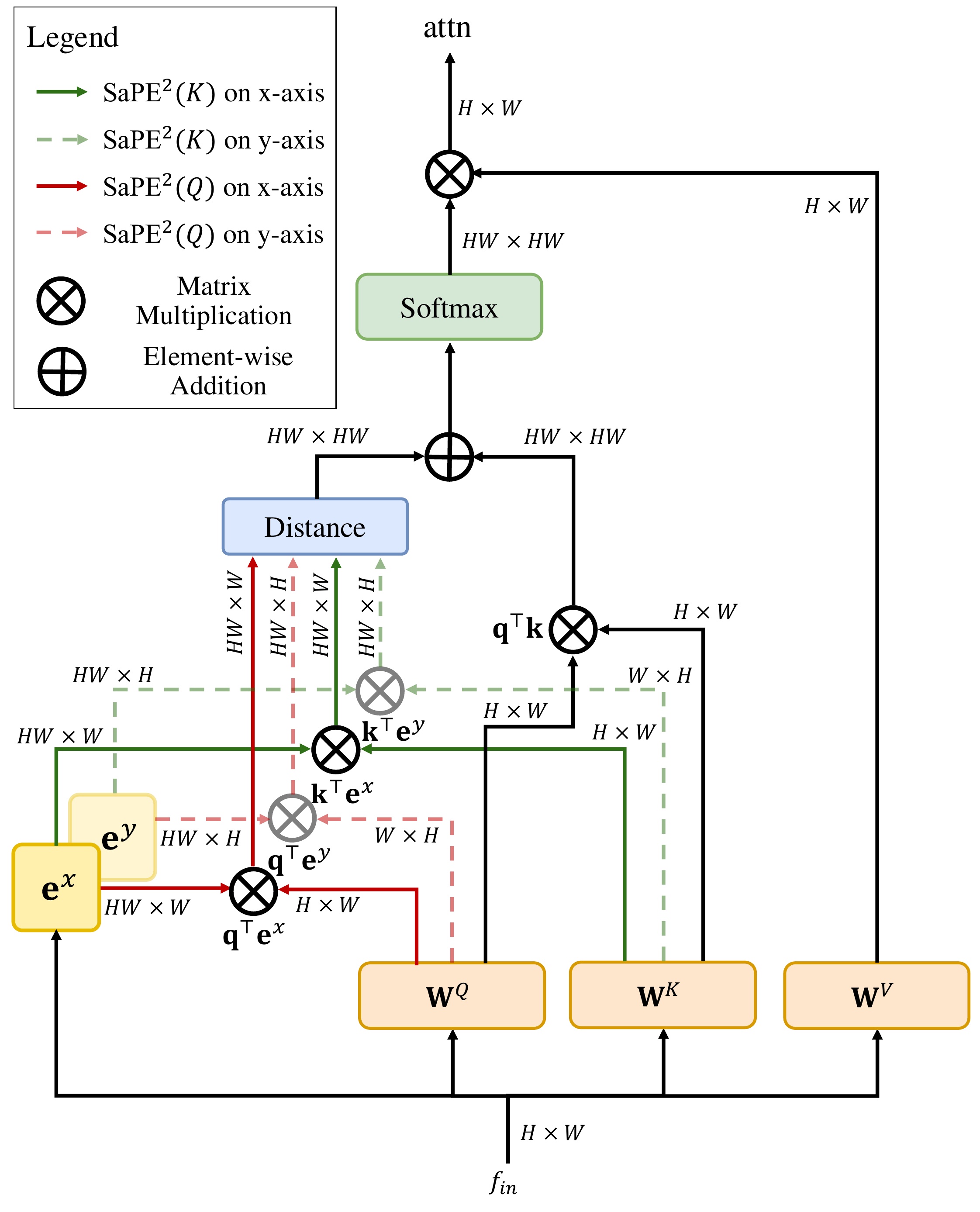}
    \caption{Self-attention mechanism with \ours, the position encoding can be applied to either key or query. For simplicity, we ignore the channel dimension here. The dimensions and symbols outside the parentheses indicate operations on the x-axis, while those inside indicate operations on the y-axis.}
    \label{fig:mode}
\end{figure}

Finally, the comprehensive position embedding that characterizes the location of the $i$-th patch along the $x$-axis is obtained through the following equation:
\begin{equation}
\label{eq:sape}
\begin{split}
    \mathbf{SaPE^2}_x \left[ i\right] &= \mathbf{q}_i^{ \top} \left[ \mathbf{e}^x\left[p_{im_1}^x\right], \mathbf{e}^x\left[p_{im_2}^x\right] \ldots \mathbf{e}^x\left[p_{im_W}^x\right]\right] 
    = \left[ z_{i1}, z_{i2}, \ldots, z_{iW} \right], \\
    &\text{ where } x_{m_1} = 1, x_{m_2} = 2, \ldots, x_{m_W} = W.
\end{split}
\end{equation}
Here $z_{ir} = \mathbf{q}_i^{\top}\mathbf{e}^x\left[p_{im_r}\right]$ answers the question: what is the contribution of the $r$-th patch's position embedding to the representation of the $i$-th patch along the $x$-axis? Except for the query mode, denoted as \ours(Q), we can also switch to the key mode $\tilde{z}_{ir} = \mathbf{k}_i^{\top}\mathbf{e}^x\left[p_{im_r}\right]$, denoted as \ours(K). The implementation details are illustrated in Figure \ref{fig:mode}. The term $z_{ir}$ serves as the attention bias added to $\mathbf{q}_i^{\top} \mathbf{k}_r$ in the 1D setting commonly used in NLP tasks. However, since we need to model the interactions between patches across both rows and columns, the neighboring content surrounding each patch must be incorporated. Therefore, the vector $\mathbf{SaPE^2}_x$ is constructed to provide a richer contextual representation by aggregating the relative position information of neighboring patches. 

In practice, directly computing and storing the vectors $\mathbf{q}_i^{ \top}\mathbf{e}^x\left[p_{im_r}\right]$ incurs additional computational and memory overhead. A more efficient approach is to first compute the multiplications $\mathbf{q}_i^{\top} \mathbf{e}^x[p]$ for all integer positions $p$, and then obtain the desired values through interpolation. The $z_i\left[p_{im_r}^x\right]$ acquired from Equation \eqref{eq:zp} is equivalent to $z_{ir}$ in Equation \eqref{eq:sape}:
\begin{equation}
    z_i[p] =\mathbf{q}_i^{\top} \mathbf{e}^x[p] \quad \text { for } p \in[0,1, \ldots, M]
\end{equation}
\begin{equation}
\label{eq:zp}
\begin{split}
    z_i\left[p_{im}^x\right]=\left(p_{im}^x-\left\lfloor p_{im}^x\right\rfloor\right) z_i\left[\left\lceil p_{im}^x\right\rceil\right]+
    \left(1-p_{im}^x+\left\lfloor p_{im}^x \right\rfloor\right) z_i\left[\left\lfloor p_{im}^x\right\rfloor\right].
\end{split}
\end{equation}

Using the $\mathbf{SaPE^2}_x$ vector, the relative positional bias between the $i$-th and the $n$-th patch along the $x$-axis can be computed based on their Euclidean distance. This formulation generalizes the bias calculation to arbitrary pairs of patches. The $n$-th patch is not constrained to share the same $y$ coordinate with the $i$-th patch, enabling the model to capture cross-row spatial interactions and enhancing its capacity to model complex spatial structures across the entire image:
\begin{equation}
    b_{i n}^x = \left\|\mathbf{SaPE^2}_x \left[ i\right] - \mathbf{SaPE^2}_x \left[ n\right]\right\|_2.
\end{equation}

By calculating the bias for the $y$-axis in the same way, we can obtain the final attention weights with \ours:
\begin{equation}
a_{im} = \frac{\mathbf{q}_i^\top \mathbf{k}_m + b_{i n}^x + b_{i n}^y}{\sqrt{d_k}}.
\end{equation}
This decomposition-based approach improves the model's ability to capture both local and global spatial dependencies. By encoding each axis separately, the model gains greater flexibility in adapting to the inherent structure of image data.

\section{Complexity Analysis}
\subsection{Space Complexity Analysis}
The space complexity of \ours~is mainly attributed to storing intermediate attention gates, positional values, and bias terms. For each axis, the attention gates require $\mathcal{O}(N \cdot W)$ storage, as they capture all pairwise interactions along a row or column, where $N=H*W$, representing the number of patches. The interpolated relative positions and their embeddings further introduce a cost of $\mathcal{O}(N \cdot W)$ and $\mathcal{O}(M \cdot d)$, respectively, where $M$ is the maximum integer position index. The interpolated position vectors $\mathbf{SaPE^2}$ occupy $\mathcal{O}(N \cdot W)$ memory, and the pairwise positional biases require $\mathcal{O}(N^2)$ space per axis. Consequently, the total space complexity is dominated by the storage of pairwise bias matrices, which scales quadratically with the number of patches. 

\subsection{Time Complexity Analysis}
The time complexity of the proposed \ours~method arises primarily from computing gated interactions and interpolated positional embeddings. For each row (or column), the gate values $g_{ij}$ are computed by evaluating inner products between all patch pairs along the same axis, resulting in a time complexity of $\mathcal{O}(N \cdot W \cdot d)$ for horizontal attention and similarly for vertical. The relative position values $p_{im}$ are accumulated through masked summations over gate values, requiring $\mathcal{O}(N \cdot W)$ operations. Interpolated embeddings are obtained by bilinear interpolation between integer positional embeddings, which introduces $\mathcal{O}(N \cdot W \cdot d)$ cost. To enhance efficiency, we adopt an optimized computation strategy that first computes dot products for all integer positions and then performs interpolation, reducing redundancy. Finally, pairwise Euclidean distances between learned positional vectors are computed using $\mathcal{O}(N^2 \cdot W)$ operations per axis, followed by attention bias calculation at $\mathcal{O}(N^2 \cdot d)$. Overall, the computational bottleneck lies in the $\mathcal{O}(N^2)$ pairwise operations.

\section{Experiments}
\label{sec:exps}

To verify the effectiveness of our proposed position encoding, we apply \ours~to the ViT model and modify the original position encoding module for training on an image classification task. 
We begin by evaluating the performance of \ours~against commonly used position encoding methods, such as APE and 2D RoPE, on image classification tasks. Next, we examine the effect of position encoding on different components of the attention mechanism (e.g., Q and K, as illustrated in Figure~\ref{fig:mode}) and its impact on downstream task performance. Finally, we empirically demonstrate that our proposed method addresses the limitations of traditional position encoding in capturing semantic information by optimizing position encoding and enhancing its capacity for semantic awareness.

    \begin{table}[t]
        \centering
        \caption{Experimental results in terms of Top 1 Accuracy (\%). Higher accuracy shows better performance.}
        \begin{tabular}{ccc}
            \toprule
            Model  & CIFAR10 & CIFAR100 \\
            \midrule
            ViT & 87.41 & 66.54 \\
            2D RoPE & 89.92 & 70.71 \\
            2D RoPE + APE & 89.98 & 71.32 \\
            CoPE + APE & 92.54 & 72.10 \\
            \midrule
            \ours & 91.06 & 71.29 \\
            \ours~+ APE & \textbf{93.98} & \textbf{72.23} \\
            \bottomrule
        \end{tabular}
        \label{tab:exp-result}
    \end{table}

\subsection{Implementation Details}
All experiments are conducted on a server equipped with four Nvidia H20 GPUs. All models are trained for 400 epochs, with a patch size of 4 and an input size of 32. The hidden dimension is set to 384, and the batch size is 128. We use the Adam optimizer to train the model. Model performance is evaluated using Top-1 Accuracy, where a higher score indicates better performance.

\subsection{Comparison between Different PE Methods}
    \label{subsec:exp-rq1}
    To demonstrate the effectiveness of our approach, we compare \ours~against commonly used position encoding methods from prior work by integrating different position encoding methods into a vanilla ViT model.
    The vanilla ViT model employs learnable absolute position encoding (APE). We replace the APE module with 2D RoPE, CoPE, and \ours~for comparison. The details of the compared methods are as follows:

        \begin{itemize}
        \item \textbf{Vanilla ViT}~\cite{vit}: We use ViT-Small as the base model, which adopts APE as its position encoding. All other comparison methods replace the position encoding in this model. We denote each method directly by the name of its position encoding module.

        \item \textbf{2D RoPE}~\cite{2d-rope}: Instead of applying the original RoPE, we directly adopt 2D RoPE as a baseline, as it introduces optimizations tailored for image data and vision models.

        \item \textbf{CoPE}~\cite{cope}: We directly apply the CoPE method from NLP to image data and vision models. CoPE serves as a generalized form of the standard relative position encoding (RPE).

        \item \textbf{*+APE}: This approach retains the original APE in ViT while incorporating additional position encoding methods. Here, APE refers to the learnable version used in the vanilla ViT.
    \end{itemize}

    Since image classification tasks require model retraining, and our goal is to efficiently evaluate the effectiveness of the position encoding module, we conduct experiments on two public datasets, CIFAR10~\footnote{https://www.cs.toronto.edu/~kriz/cifar-10-python.tar.gz} and CIFAR1100~\footnote{https://www.cs.toronto.edu/~kriz/cifar-100-python.tar.gz}. 

    \begin{table}[t]
        \centering
        \caption{Experimental results of image classification on CIFAR10 with APE and different settings of \ours.}
        \begin{tabular}{ccc}
            \toprule
            & Top 1 Acc.(\%) & Top 5 Acc.(\%) \\
            \midrule
            APE & 87.41 & 99.47 \\
            \ours(K) + APE & \textbf{93.98} & \textbf{99.71} \\
            \ours(Q) + APE & 90.17 & 99.46 \\
            \bottomrule
        \end{tabular}

        \label{tab:exp-discussion}
    \end{table}

    \begin{figure*}
        \centering
        \includegraphics[width=0.9\linewidth]{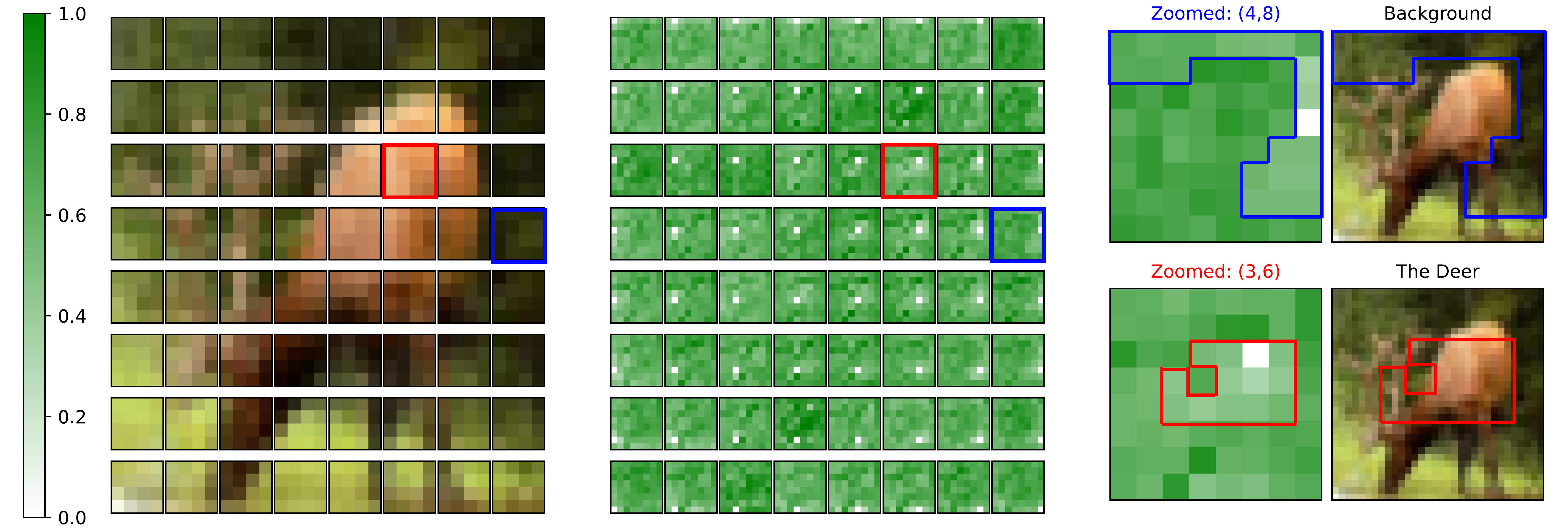}\\
        \includegraphics[width=0.9\linewidth]{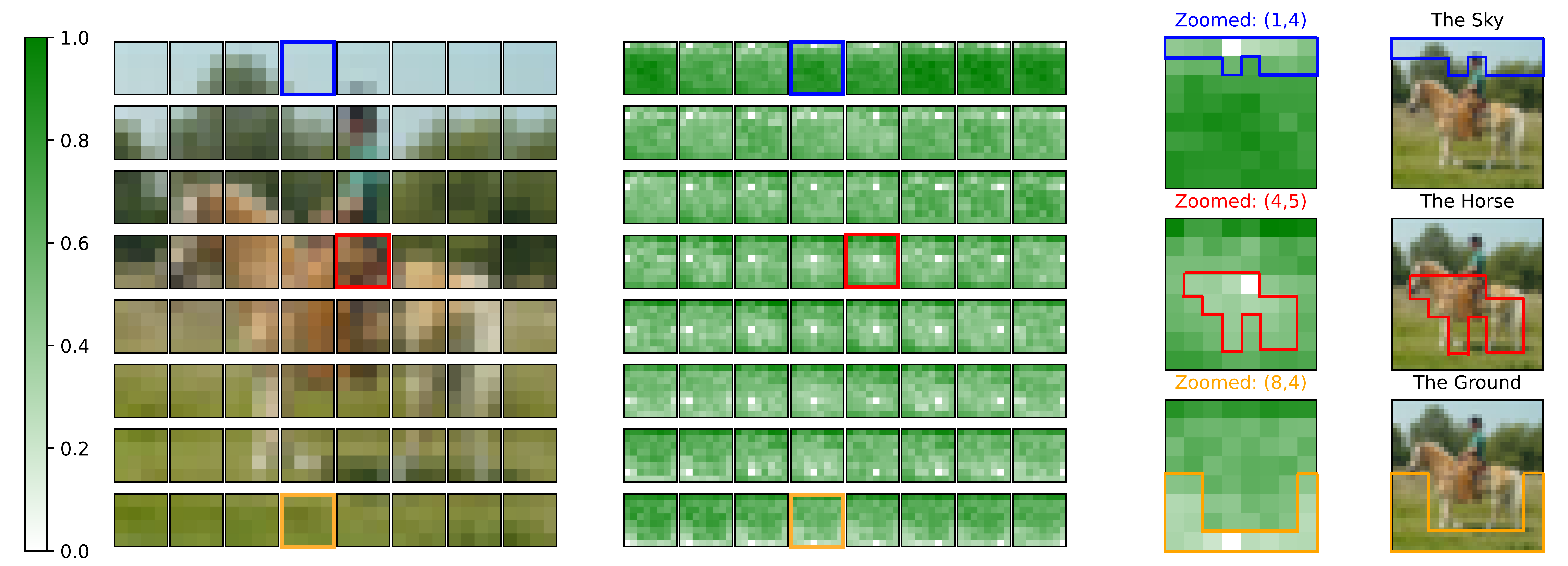}
        \caption{Visualization of the attention bias calculated based on \ours. The patch size is set to 4, resulting in 8×8 patches for 32×32 images. The color transition from white to green indicates further semantic distance. \textbf{Left}: the original image; \textbf{Middle}: the \ours~value between different patches; \textbf{Right}: visualization of \ours~on patch from the background and the subject.The results indicate that our position encoding method effectively helps the model perceive contextual information.}
    \label{fig:exp-case}
    \end{figure*}

    The experimental results are presented in Table~\ref{tab:exp-result}.

    As observed, \ours~achieves the best performance. We further provide the following analysis:

    (1) Incorporating semantic information of patches into position encoding enhances model performance. Compared to APE and 2D RoPE, strategies that incorporate semantic information, such as CoPE and \ours, enable the model to better differentiate instances of similar objects, thereby achieving superior performance.
    
    (2) The 2D version of position encodings enhance the model’s ability to capture spatial structures, thereby improving prediction accuracy. The comparison of RoPE and 2D RoPE, CoPE+APE and \ours+APE demonstrates the effectiveness of integrating 2D spatial information, enabling the model to better capture spatial relationships within the input image.

    (3) Integrating both absolute and relative position information provides additional positional cues to the model, thereby improving its performance. This suggests that absolute and relative encodings provide complementary information: while relative encoding captures positional relationships, absolute encoding offers explicit positional differentiation, enhancing the model’s ability to understand sequence structure.

\subsection{Analysis of Integrating PE with Attention}

    As discussed in the methodology section, our proposed \ours~can be integrated into the query (Q) and the key (K) within the self-attention module. To assess the effectiveness of applying \ours~to different components of the attention module in vision transformers, we design three experimental settings: applying \ours~to Q or K, or both Q and K simultaneously, denoted as \ours(Q), \ours(K). All other experimental settings follow the description in Section~\ref{subsec:exp-rq1}.

    The results are presented in Table~\ref{tab:exp-discussion}. We analyze the results as follows:

    (1) Compared to solely using APE (i.e., the default PE setting in ViT), \ours~significantly enhances model performance (Table~\ref{tab:exp-discussion} Line 1 vs. Lines 2 and 3). This demonstrates that semantic-aware position encoding enables the model to more effectively capture both semantic and spatial relationships.

    (2) Integrating \ours~with K yields the best performance, indicating that improving the query’s semantic and spatial awareness within the attention module is the most effective approach. This is because, during attention computation, enriching K with semantic-aware PE enables the query to more effectively capture spatial relationships within the context. As a result, the model exhibits an enhanced ability to capture spatial dependencies among patches, leading to improved performance.

\subsection{Case Study}
    To further validate the effectiveness of our proposed PE strategy, we visualize the attention bias induced by \ours~in ViT-Small on the CIFAR-10 validation set. Figure~\ref{fig:exp-case} presents two examples. 
    The images are divided into 8×8 patches, and each patch is represented as a small block in the attention map, illustrating its attention bias with respect to other patches.

    Effectively recognizing and distinguishing the subject and background plays a crucial role in computer vision tasks, contributing to improved prediction performance.  
    Take the ``horse'' image (Figure~\ref{fig:exp-case} bottom) as an example.
    The image can generally be divided into three parts: the sky above, the human and horse subjects in the center, and the ground below. Correspondingly, the PE maps on the right reveal that patches in the top two rows exhibit smaller distance from patches in the same region (i.e., the sky) while distant from the human, the horse, and the ground below. Similar patterns are observed for the subject and the ground. This indicates that patches tend to focus more on others with similar semantic contexts.

    We attribute this effective attention mechanism to our proposed semantic-aware position encoding, \ours, which enables ViT to better capture positional relationships between patches based on semantic similarity, thereby enhancing its performance in vision tasks.

\section{Conclusion}
\label{sec:conclusion}
In this paper, we propose a novel 2-dimensional semantic-aware position encoding method, \ours, to address the limitations of existing position encoding techniques in transformer-based vision models, which primarily focus on absolute or relative spatial coordinates while overlooking relationships between semantically similar yet spatially distant regions. To this end, \ours~dynamically adapts position representations based on contextual information, effectively capturing semantic relationships between different patches. Experimental results demonstrate that by bridging the gap between position encoding and perceptual similarity, \ours~significantly enhances the performance of transformer-based vision models, presenting a promising direction for future research.
The code of our proposed method will be open-sourced upon publication.

\clearpage
\newpage
\small
\bibliographystyle{unsrt}
\bibliography{reference}

\newpage
\appendix

\section{Related Work}
\label{sec:related_work}

\subsection{Position Encodings in NLP}
Position encoding was first introduced in NLP transformers, where absolute position encoding (APE) \cite{transformer, gehring2017convolutional} and relative position encoding (RPE) \cite{shaw2018self, raffel2020exploring} became standard approaches. APE assigns fixed or learned embeddings to tokens based on absolute positions, often using sinusoidal functions or trainable vectors. However, APE does not capture relative positional information, making it sensitive to sequence length variations.

To address this, relative position encoding (RPE) modifies the attention mechanism to incorporate relative position biases, improving generalization to different sequence lengths. Rotary Position Embedding (RoPE) \cite{rope} extends this by encoding relative angular displacement through rotation matrices, which enhances the model’s ability to handle longer sequences effectively. 

Contextual position encoding (CoPE) \cite{cope} extends the idea of position encoding by dynamically adjusting positional representations based on contextual information rather than relying solely on predefined spatial coordinates. Unlike traditional absolute or relative position encodings, CoPE modulates position embeddings based on local content, allowing the model to better capture hierarchical and structural dependencies within a sequence. This adaptability enables improved generalization in scenarios where positional importance varies dynamically, such as in structured sequences or tasks requiring position-dependent reasoning.

\subsection{Vision Transformers}
The impact of Transformer models has expanded beyond natural language processing to various domains, including computer vision. Vision Transformer (ViT) \cite{vit} is the first model in the field of computer vision to adopt a Transformer-based architecture as its backbone. ViT partitions an input image into multiple fixed-size patches, which are treated as tokens in a sequence and processed through a Transformer encoder. This approach has demonstrated superior performance compared to traditional CNN-based models in image classification tasks. Subsequent models have introduced various modifications to further enhance performance based on the ViT framework. T2T-ViT \cite{yuan2021tokens} enhances the tokenization process by progressively aggregating neighboring tokens, thereby capturing local structure information more effectively. DeiT \cite{touvron2021training} introduces a distillation token during training, enabling the model to learn from a teacher network and improving data efficiency. Pyramid ViT \cite{wang2021pyramid} and PiT \cite{heo2021rethinking} employ hierarchical architectures, facilitating multi-scale feature representation and reducing computational complexity. CvT \cite{wu2021cvt} and CeiT \cite{yuan2021incorporating} integrate convolutional layers into the Transformer framework, providing inductive biases that enhance the modeling of local features. Swin Transformer \cite{swin} utilizes a shifted window-based attention mechanism, which limits self-attention computation to non-overlapping local windows and achieves efficient modeling of long-range dependencies.

Transformer were later extended to downstream tasks in computer vision, including object detection and semantic segmentation. DETR \cite{detr} introduced an end-to-end object detection framework using a CNN backbone with a Transformer encoder-decoder, inspiring further optimizations \cite{zhu2020deformable, zheng2020end, gao2021fast}. Several models have employed vanilla ViT as a backbone, such as ViTDet \cite{li2022exploring}, which incorporates upsampling and downsampling; and ViT-Adapter \cite{chenvision}, which adds inductive biases for detection and segmentation tasks. SETR \cite{zheng2021rethinking} first adopted ViT with a CNN decoder for semantic segmentation, while Segmenter \cite{strudel2021segmenter} introduced a Transformer decoder. SegFormer \cite{xie2021segformer} employed a hierarchical encoder to improve segmentation performance.

\subsection{Position Encodings in CV}
Previous transformer-based vision models often adopted absolute position encoding, relative position encoding, or rotary position encoding, similar to methods used in natural language processing. Given that vision models inherently process 2D inputs, some subsequent models have adapted these NLP position encodings to two-dimensional formats.

A straightforward and simple idea is to extend relative position encoding to a 2D formulation \cite{wu2021rethinking}. This involves computing relative position encodings separately for the horizontal (x-axis) and vertical (y-axis) directions. These directional encodings are then combined to effectively represent the 2D relative positional relationships between pixels, enhancing the model’s ability to understand spatial hierarchies and patterns in visual data. Furthermore, 2D RoPE \cite{2d-rope} generalizes RoPE \cite{rope} to a two-dimensional setting. It achieve this by implementing axial and mixed learnable frequency approaches. The axial frequency method applies RoPE separately to the x-axis and y-axis, while the mixed learnable frequency approach introduces learnable parameters to handle diagonal directions, aiming to capture more complex spatial relationships in images. 

Following works further improve the position encoding for CV in different aspects. A representative work is LaPE \cite{yu2023lape}. It identify that PEs are added to patch embeddings before entering the Transformer encoders, with both undergoing the same layer normalization. This shared normalization can restrict the expressiveness of PEs, potentially hindering model performance. LaPE involves assigning separate layer normalizations to token embeddings and PEs within each Transformer layer, allowing each to maintain its unique characteristics. Additionally, PEs are progressively adapted and delivered across layers, providing layer-specific positional information and enabling the model to adjust to varying levels of abstraction throughout the network.

\section{Datasets}
\label{app:data}
We conduct experiments on public datasets including CIFAR10~\footnote{https://www.cs.toronto.edu/ kriz/cifar-10-python.tar.gz} and CIFAR100~\footnote{https://www.cs.toronto.edu/ kriz/cifar-100-python.tar.gz}. Detailed statistics are shown in Tab.~\ref{tab:appendix-data}.

\begin{table}[h]
    \centering
    \caption{Detailed statistics, including image size, categories, and data splits.}
    \begin{tabular}{ccccc}
        \toprule
        Dataset & size & categories & training size & eval size\\
        \midrule
        CIFAR10 & 32 & 10 & 50,000 & 10,000 \\
        CIFAR100 & 32 & 100 & 50,000 & 10,000 \\
        \bottomrule
    \end{tabular}

    \label{tab:appendix-data}
\end{table}

\section{Limatations and Future Work}
\label{app:limitation}
While our method demonstrates strong performance, a current limitation lies in its limited generalizability across different backbone architectures. Specifically, when the vision transformer backbone is changed, it may be necessary to re-pretrain the model with our proposed \ours~as the positional embedding, which introduces additional computational cost. Nevertheless, this also opens up a promising research direction. In future work, we plan to investigate fine-tuning strategies to improve the adaptability of \ours~and reduce the need for full retraining across diverse architectures.

\section{Codes Implementation}
\label{app:code}
Below is the core implementation code of \ours, which can be equipped within the transformer's attention calculation module. To use it, you can integrate \ours~into the transformer's attention layer. Specifically, this can be achieved by adding the output of SaPE to the attention computation.
\lstset{
    backgroundcolor=\color[RGB]{245,245,244},
    columns=fullflexible,
    linewidth=1\linewidth, 
    breaklines=true,
    keywordstyle=\bfseries\color{NavyBlue},
    commentstyle=\it\color[RGB]{0,96,96},
    language=Python}
\begin{lstlisting}
class SaPE_unit(nn.Module):
    def __init__(self, npos_max, head_dim, scale):
        super(SaPE_unit, self).__init__()
        self.npos_max = npos_max
        self.pos_emb = nn.Parameter(torch.zeros(1, head_dim, npos_max))
        self.scale = scale
        trunc_normal_(self.pos_emb, std=.02)

    def forward(self, q, k, v):
        # query: (batch_size, heads, seq_len, head_dim)
        # attn_logits: (batch_size, heads, seq_len, seq_len)
        # compute positions
        
        attn_logits = (q * self.scale) @ k.transpose(-2, -1)
        gates = torch.sigmoid(attn_logits)
        
        pos = gates.flip(-1).cumsum(dim=-1).flip(-1)
        pos = pos.clamp(max=self.npos_max - 1)
        # interpolate from integer positions
        pos_ceil = pos.ceil()
        pos_floor = pos.floor()
        logits_int = torch.matmul(q, self.pos_emb)
        logits_ceil = logits_int.gather(-1, pos_ceil)
        logits_floor = logits_int.gather(-1, pos_floor)
        w = pos - pos_floor
        return logits_ceil * w + logits_floor * (1 - w)
    
    
class SaPE(nn.Module):
    def __init__(self, npos_max, head_dim, scale, num_heads, num_patches, img_size):
        super(SaPE, self).__init__()
        self.npos_max = npos_max
        self.sape_x = SaPE_unit(npos_max, head_dim, scale)
        self.sape_y = SaPE_unit(npos_max, head_dim, scale)

    def forward(self, query, key, value):
        # query: (batch_size, heads, seq_len, head_dim)
        
        B, heads, seq_len, head_dim = query.shape
        W = H = int(math.sqrt(seq_len))
        query = query.reshape(B, heads, H, W, head_dim)
        key = key.reshape(B, heads, H, W, head_dim)
        value = value.reshape(B, heads, H, W, head_dim)
        # [batch, heads, patch_h, patch_w, hidden], [batch, h/2, w/2]

        # split q, k, v
        q_x = query.permute(0, 2, 1, 3, 4).reshape(B*H, heads, W, head_dim)
        q_y = query.permute(0, 3, 1, 2, 4).reshape(B*W, heads, H, head_dim)

        k_x = key.permute(0, 2, 1, 3, 4).reshape(B*H, heads, W, head_dim)
        k_y = key.permute(0, 3, 1, 2, 4).reshape(B*W, heads, H, head_dim)
        
        v_x = value.permute(0, 2, 1, 3, 4).reshape(B*H, heads, W, head_dim)
        v_y = value.permute(0, 3, 1, 2, 4).reshape(B*W, heads, H, head_dim)
        
        w_x = self.sape_x(q_x, k_x, v_x).reshape(B, H, heads, W, H)
        w_y = self.sape_y(q_y, k_y, v_y).reshape(B, W, heads, H, W)
        
        w_x = w_x.permute(0, 2, 1, 3, 4).reshape(B * heads, H * W, H)
        w_y = w_y.permute(0, 2, 3, 1, 4).reshape(B * heads, H * W, H)

        # Calculate the 2-dimentional relative distance
        pe_res_x = torch.cdist(w_x, w_x, p=2).reshape(B, heads, H*W, H*W)
        pe_res_y = torch.cdist(w_y, w_y, p=2).reshape(B, heads, H*W, H*W)
        
        return pe_res_x + pe_res_y

\end{lstlisting}


\end{document}